\documentclass[cameraready]{Interspeech}

\title{HybridCodec: Modeling Discrete and Continuous Representations\\For Efficient Speech Language Models}

\author[affiliation={1,2}, equalcontribution]{Artem}{Ploujnikov}
\author[affiliation={3}, equalcontribution]{Francesco}{Verdini}
\author[affiliation={4}]{Samir}{Sadok}
\author[affiliation={1,2}]{Mirco}{Ravanelli}

\address{
    $^1$Mila, Quebec AI Institute, Canada; 
    $^2$Concordia University, Canada; \\
    $^3$Sapienza University of Rome, Italy;
    $^4$Inria, Université Grenoble Alpes CNRS, LJK, France 
}

\email{artem.ploujnikov@mail.concordia.ca, francesco.verdini@uniroma1.it, \\samir.sadok@inria.fr, mirco.ravanelli@mail.concordia.ca}


\usepackage{graphicx}
\usepackage{comment}
\usepackage{tabularx}
\usepackage{multirow}
\usepackage[table]{xcolor}
\usepackage{amsmath}
\usepackage{amsfonts}
\usepackage{amssymb}
\usepackage{todonotes}
\usepackage{booktabs}
\usepackage{float}
\usepackage{subcaption}
\usepackage{pifont}
\usepackage{wrapfig}
\usepackage{caption}
\usepackage{comment}
\usepackage{cite}
\newcommand{\hide}[1]{}

\newif\ifresynthft
\resynthfttrue

\begin{document}

\maketitle

\begin{abstract}
Discrete audio representations have become increasingly popular for building multimodal text-audio systems and integrating audio capabilities into Large Language Models (LLMs). However, numerous studies report performance degradation on various downstream tasks due to information loss during discretization. To address this, we propose a novel approach combining temporally compressed discrete tokens with dimensionality-reduced continuous residuals. Our framework consists of a hybridized discrete-continuous focal modulation codec and a hybrid Transformer. This architecture performs autoregressive inference in the discrete domain, coupled with non-autoregressive prediction and continuous residual upsampling. Experimental results show that our approach significantly improves the retention of speaker characteristics compared to discrete-only methods, while simultaneously reducing the  number of required autoregressive steps.
\end{abstract}

\section{Introduction}
\label{sec:introduction}

The human mind processes the world through a complex interplay of discrete categories and continuous spectra \cite{discrete-continuous-brain, attractor-integrator}. Human language perfectly illustrates this duality. It imposes a clear \emph{discrete} hierarchy (sequences of phonemes forming words and sentences captured in alphabets or logographic systems) onto a rich modulation of \emph{continuous} characteristics, such as pitch, tone, emotion, and prosody.  

The advent of the Transformer architecture \cite{transformer} established discrete token sequences as the \textit{de facto} medium for modern artificial intelligence. This paradigm, which drives autoregressive generation and  Large Language Models (LLMs)~\cite{gpt, llama, gemini}, was subsequently adapted for the audio domain. Pioneering architectures like the Vector Quantized Variational Autoencoder (VQ-VAE) \cite{vqvae} demonstrated that continuous information can be effectively compressed into a discrete latent space, motivating the development of neural audio codecs (NACs) \cite{dates, kyutai2024moshi, xin2024bigcodec, dac}.  Fundamentally, a NAC comprises an encoder, a vector quantizer, and a decoder that map continuous audio into low-bitrate discrete tokens and back to the waveform. Unlike traditional codecs such as MP3 that rely on algorithmic signal processing and psychoacoustics, NACs learn a finite or variable data-driven \emph{vocabulary} of sounds. This allows them to achieve extreme compression rates while preserving rich semantic and acoustic features, effectively bridging the gap between raw signal processing and natural language modeling by enabling LLMs to process speech as natively as text. Models like AudioLM (semantic and acoustic modeling)~\cite{borsos2023audiolm}, VALL-E (zero-shot voice cloning)~\cite{valle}, and SpeechGPT (cross-model speech-text LLM)~\cite{speechgpt} have successfully leveraged these discrete audio tokens to drive significant breakthroughs in zero-shot speech synthesis and end-to-end multimodal dialogue.

Despite their advantages, fully discrete representations introduce an inherent quantization penalty. As evidenced by Benchmarks (e.g., SUPERB \cite{superb}, DASB~\cite{dasb-benchmark}) and recent comparative surveys \cite{dates, speechdt, kammoun2025modeling} highlight a fundamental trade-off: while discrete tokens facilitate stable convergence and seamless LLM integration, the quantization process irreversibly discards fine-grained acoustic details. Fundamentally, this loss stems from the classic rate-distortion trade-off \cite{cover1999elements, shannon1959coding}. At low bit-rates, NACs prioritize semantic content intelligibility over acoustic richness, lacking the bandwidth to encode micro-prosody and speaker timbre \cite{dates}. To mitigate this limitation, we propose a novel \emph{hybrid} paradigm in which the codec supports optional refinements through high-frame-rate continuous residuals, and an LM can start with a lossy, low-resolution approximation and then compute a one-step continuous refinement, vastly reducing the total number of forward passes required in inference. 

Our main contributions are as follows:
(1) \textbf{HybridCodec}, a novel NAC framework extending FocalCodec~\cite{focalcodec, focalcodec-streaming}, which jointly extracts time-reduced discrete tokens and models the remaining information as dimensionality-reduced continuous residuals;
(2) \textbf{HybridLM}, a decoder-only Transformer~\cite{transformer} designed to process these hybrid representations. It unifies efficient, low-frame rate autoregressive (AR) prediction for discrete tokens with a single-step non-autoregressive (NAR) prediction and continuous residual upsampling;
(3) A unified framework that leverages the HybridLM architecture to effectively handle major downstream speech tasks, including ASR and TTS, within a single framework.
\begin{figure*}[t!]
    \centering
    \includegraphics[width=\linewidth]{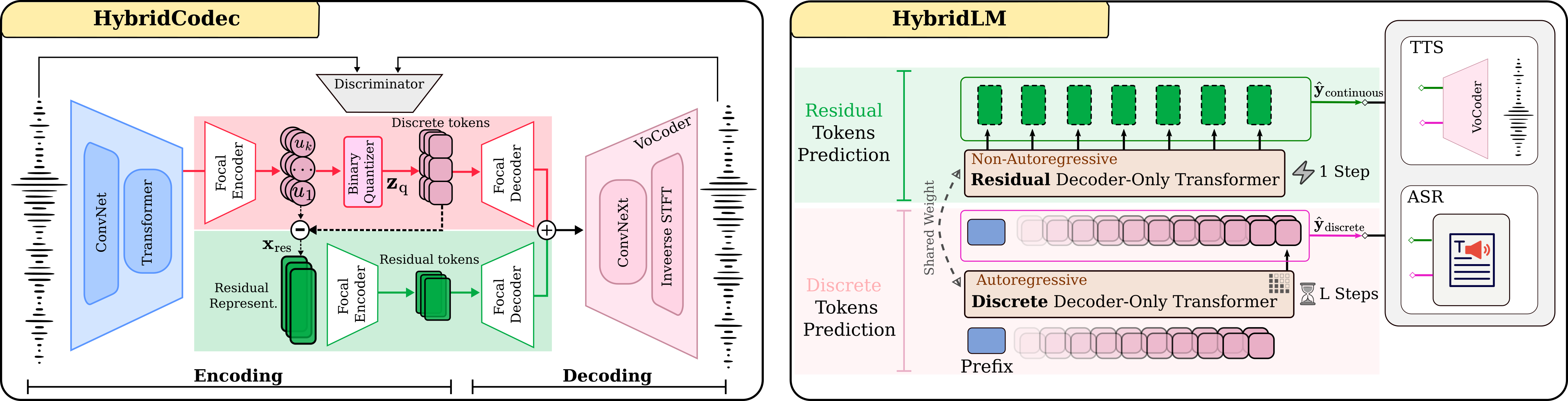}
    \caption{Overview of the proposed architecture: \textbf{HybridCodec} (left) provides dual-path discrete-continuous compression, and \textbf{HybridLM} (right) unifies these representations through interleaved autoregressive and non-autoregressive decoding.}
    \label{fig:arch}
\end{figure*}
This hybrid paradigm restores the fine-grained information lost in discrete LMs. Experimental results on LibriTTS \cite{libritts} dataset showed that our approach significantly outperforms discrete baselines, especially at extremely low frame rates such as 6.25\,Hz, while substantially reducing AR steps.

\section{Related Work}
 Recent work has been responding to the discrete-continuous performance gap with further task-specific analysis and a variety of adaptations. Studies in ASR \cite{discrete-continuous-asr} confirm the gap, showing that such an information bottleneck directly degrades downstream performance by stripping the signal of its prosodic nuance and speaker identity. To overcome this limitation, recent literature has explored re-integrating continuous features through diffusion mechanisms, continuous autoregressive modeling, or masked modeling \cite{clear-tts, spear-tts, sadok2026residual}. However, these approaches remain highly task-specific, sacrificing the unified, generalizable framework that discrete LLMs provide. This limits their ability to handle diverse speech applications (like generation and recognition) within a single model. Discrete-continuous hybridization has been successfully explored in other domains, such as RL and robotics\cite{hyar, discrete-continuous-em, discrete-continuous-robot}, text diffusion\cite{discrete-continuous-diffusion}, and others.
This raises a critical question: \textit{is it possible to design a unified language model that leverages the efficiency of discrete tokens while restoring the rich acoustic nuances of continuous speech?}
To the best of our knowledge, our approach is the first to unify discrete and continuous refinement within a single Transformer architecture. By leveraging these two domains, we achieve high-fidelity speech synthesis at ultra-low frame rates.

\section{Model Architecture}
\subsection{Preliminaries: The FocalCodec Architecture}
FocalCodec~\cite{focalcodec} employs an asymmetric VQ-VAE architecture centered around a compressor-quantizer-decompressor bottleneck. It uses the first six layers of a pretrained WavLM as a base encoder to extract jointly acoustic and semantic features. Its core pipeline relies on \emph{focal modulation}: a \emph{Focal Encoder (FE)} (compressor) downsamples these continuous features into a compact latent space in linear time by aggregating multi-scale global and local contexts (noted as $\mathbf{x}_{\text{base}}$). The representations are then discretized using Binary Spherical Quantization (BSQ) \cite{zhao2024image}, a lookup-free approach that enforces bounded quantization errors and maximizes codebook utilization. Then, a \emph{Focal Decoder (FD)} (decompressor) mirrors the downscaling process to upsample the discrete tokens, and explicitly reconstruct the original continuous WavLM representations. Finally, a lightweight Vocos decoder~\cite{vocos} synthesizes the audio waveform directly from these restored continuous features.

\subsection{HybridCodec: Extracting Hybrid Representations}
\label{sec:hybridcodec}
The HybridCodec, shown in Figure \ref{fig:arch} (left), extends FocalCodec~\cite{focalcodec} by adding a secondary pathway. This branch, consisting of an additional focal encoder and decoder, captures and compresses the continuous residual information lost during discretization.

\noindent\textbf{Encoding: Dual-Path Feature Extraction.~} 
The encoding process maps the base representations, $\mathbf{x}_{\text{base}} \in \mathbb{R}^{T \times d_{\text{}}}$, into a dual discrete-continuous latent space. 
First, the \emph{discrete pathway} (highlighted in red in Fig.~\ref{fig:arch}) extracts the quantized indices $\mathbf{z}_{q} = \mathrm{FQ}_\theta(\mathbf{x}_{\text{base}})$. From these indices, we derive the quantized approximation $\hat{\mathbf{x}}_{\text{quant}} = \mathrm{BSQ}_\theta^{-1}(\mathbf{z}_{q})$.
Second, the \emph{continuous pathway} (highlighted in green in Fig.~\ref{fig:arch}) captures the fine-grained acoustic details lost to quantization by computing the residual error: $\mathbf{x}_{\text{res}} = \mathbf{x}_{\text{base}} - \hat{\mathbf{x}}_{\text{quant}}$. This continuous residual is compressed by a dedicated residual focal encoder, $\mathrm{FE}_{\text{res}}$, which applies a temporal down-sampling stride $r$ to yield a dimensionality-reduced bottleneck representation: $\bar{\mathbf{x}}_{\text{res}} = \mathrm{FE}_{\text{res}}(\mathbf{x}_{\text{res}})$. 
To control the temporal resolution, we adjust the strides of $\mathrm{FE}_{\text{res}}$: $(1, 1, 1)$ for $50~\text{Hz}$, $(2, 1, 1)$ for $25~\text{Hz}$, $(2, 2, 1)$ for $12.5~\text{Hz}$, and $(2, 2, 2)$ for $6.25~\text{Hz}$.

\noindent\textbf{Decoding: Feature Fusion and Reconstruction.~}
The decoding process perfectly mirrors the encoding stages to reconstruct the full hybrid signal. 
First, the \emph{discrete pathway} projects the indices $\mathbf{z}_{q}$ back into the continuous embedding space via the inverse quantizer: $\hat{\mathbf{x}}_{\text{quant}} = \mathrm{FQ}_\theta^{-1}(\mathbf{z}_{q})$.
Second, the \emph{continuous pathway} passes the bottleneck residual $\bar{\mathbf{x}}_{\text{res}}$ through a residual focal decoder, $\mathrm{FD}_{\text{res}}$. This module upsamples the representation by the factor $r$ to restore the original temporal resolution: $\hat{\mathbf{x}}_{\text{res}} = \mathrm{FD}_\text{res}(\bar{\mathbf{x}}_\text{res}) \in \mathbb{R}^{T\times d_\text{}}$.
Finally, the full representation is synthesized by adding both streams together before passing them to the Vocos decoder: $\hat{\mathbf{x}}_{\text{base}} = \hat{\mathbf{x}}_{\text{quant}} + \hat{\mathbf{x}}_{\text{res}}$.

\begin{table*}
    \footnotesize
    \centering
    \caption{
    Resynthesis performance between baseline codecs and our hybrid codec. $\uparrow/\downarrow$ indicates higher/lower is better. \textbf{bold} and \underline{second} denote the best and second-best results, respectively.
    }
\begin{tabular}{ccccccc}
\toprule
 \multirow{2}{*}{\textbf{NAC}}   & \multirow{2}{*}{\textbf{Frame rate}}    & \multirow{2}{*}{\textbf{UTMOS} ($\uparrow$)} &   \multirow{2}{*}{\textbf{dWER} ($\downarrow$)} & \multirow{2}{*}{\textbf{SpkSim} ($\uparrow$)} &  \multirow{2}{*}{\textbf{Code Usage} ($\uparrow$)} &  \multirow{2}{*}{\textbf{Norm Entropy} ($\uparrow$)} \\
 \\
\hline
Reference & --- & 4.09 & 0.00 & 100.0 & --- & ---   \\
DAC \cite{kumar2023high} & 50\,Hz & 1.29 & 20.04 & 89.2 & \textbf{100.0} & 91.7  \\
Mimi \cite{kyutai2024moshi} & 12.5\,Hz & 3.29 & 5.73 & 96.0 & 95.6 & 91.8  \\
BigCodec \cite{xin2024bigcodec} & 50\,Hz & 4.11 & 2.55 & \textbf{ 98.5} & \textbf{100.0} & \textbf{98.6}      \\
FocalCodec \cite{focalcodec} & 12.5\,Hz & \textbf{4.22}  & 7.94 & 93.9 & 98.2 & 97.4        \\
FocalCodec \cite{focalcodec} & 25\,Hz  &  \underline{4.14} & 3.30 & 96.3 & 99.8 & \underline{98.4}        \\
 \hline
HybridCodec & 50\,Hz & 4.07  & \textbf{1.47} & \underline{97.2} &\underline{99.9} & 96.3   \\
HybridCodec & 25\,Hz & 4.07  & \underline{1.48} & 96.7 & 98.8 & 96.8  \\
HybridCodec & 12.5\,Hz & 4.09 &\textbf{1.47} & 96.2 & 97.1 & 96.7  \\
HybridCodec & 6.25\,Hz  &  3.98 & 1.50 & 97.1     & 97.4 & 98.2  \\
\bottomrule
\end{tabular}
    \label{tab:results-resynth-wavlm}
\end{table*}

\subsection{HybridLM Architecture}
HybridLM is a GPT-style~\cite{gpt} decoder-only Transformer, illustrated in Figure~\ref{fig:arch} (right), tailored to process the dual representations of HybridCodec (Section~\ref{sec:hybridcodec}). It unifies autoregressive (AR) and non-autoregressive (NAR) decoding within a single network: discrete tokens drive the AR phase to establish semantic structure, while continuous residuals are predicted in a NAR pass to recover high-fidelity acoustic details. Unlike VALL-E~\cite{valle}, our model supports mixed discrete-continuous prompts at different temporal scales. The model was designed to fully exploit HybridCodec features, including both semantic indices (in AR mode) and continuous residuals (single-pass NAR).

\noindent\textbf{Unified AR and NAR Modeling via AdaLN.} 
Combining AR classification (token generation) and NAR regression (residual prediction) risks objective interference in deeper layers if relying on simple prefix conditioning. To mitigate this, we employ Adaptive Layer Normalization (AdaLN) to multiplex both operational modes. By injecting a mode-specific embedding ($i_{\text{mode}} \in \{\text{AR}, \text{NAR}\}$) at every layer, AdaLN provides deep conditioning that dynamically adapts internal representations. This effectively creates two specialized, interference-free sub-models within a shared backbone~\cite{adaspeech, valle}. We train models with 12 layers, 4 attention heads, $d_\textrm{model} = d_\textrm{emb} = 512$, and $d_\textrm{ffn} = 2048$ (inner dimension of the feed-forward layers).

Given a decoding mode identifier $i_{\text{mode}}$, the AdaLN modulation parameters are computed as follows:
\begin{align*}
\mathbf{e} &= \mathrm{Emb}(i_\text{mode}) && \text{(\footnotesize Mode embedding)} \\
\boldsymbol{\gamma} &= \mathbf{W}_\gamma \mathbf{e} + \mathbf{b}_\gamma && \text{(\footnotesize Scaling vector)} \\
\boldsymbol{\beta} &= \mathbf{W}_\beta \mathbf{e} + \mathbf{b}_\beta && \text{(\footnotesize Bias vector)} \\
\bar{\mathbf{z}} &= \mathrm{LayerNorm}(\mathbf{z}) && \text{(\footnotesize Standard LN)} \\
\mathbf{z}_\text{cond} &= \boldsymbol{\gamma} \odot \bar{\mathbf{z}} + \boldsymbol{\beta} && \text{(\footnotesize Affine transform)}
\end{align*}
where $\mathrm{Emb}(\cdot)$ is a learned embedding layer mapping the discrete mode identifier to a continuous vector $\mathbf{e}$, $\mathbf{W}_\gamma$ and $\mathbf{W}_\beta$ are learnable weight matrices, $\mathbf{b}_\gamma$ and $\mathbf{b}_\beta$ are bias terms, $\vec{z}$ denotes a latent variable (usually the output of the previous layer or of attention/FFN) and $\odot$ denotes the Hadamard product.

\noindent\textbf{Speaker Embeddings.} To condition the generation on a specific voice, we inject pretrained ECAPA-TDNN~\cite{ecapa-tdnn} speaker embeddings, extracted using the SpeechBrain~\cite{speechbrain_v1} toolkit. These embeddings are integrated via a simple linear projection and addition to all token embeddings in the source sequence.   

\noindent\textbf{Training Procedure.} 
Both the discrete and continuous paths are trained conventionally with teacher forcing, as in the original Transformer~\cite{transformer} and the two losses (NLL for discrete and MSE for continuous) are combined.

\noindent\textbf{Cascaded Inference.} During the inference phase, the generation proceeds in a cascaded manner. Given a task-specific conditioning sequence $\mathbf{c}$ (e.g. text, phonemes or an acoustic prefix), the discrete tokens are first generated autoregressively. Subsequently, the continuous residuals are predicted in a single non-autoregressive forward pass:
\begin{align*}
\hat{\mathbf{z}}_{q} &= \mathrm{AR}(\mathbf{c}) && \text{(\footnotesize AR generation)} \\
\mathbf{h}_{\text{NAR}} &= \big[\mathbf{c} \parallel \mathrm{Up}(\hat{\mathbf{z}}_{q}, r)\big] && \text{\footnotesize (Upsample \& Concat)}\\
\hat{\mathbf{z}}_{\text{res}} &= f_\textrm{SLT}^{-1}(\mathrm{NAR}(\mathbf{h}_{\text{NAR}})) && \text{(\footnotesize NAR prediction)}\\
\hat{\mathbf{s}} &= \mathrm{Decoder}(\hat{\mathbf{z}}_{q}, \hat{\mathbf{z}}_{\text{res}}) && \text{(\footnotesize Waveform synthesis)}
\end{align*}
where $\mathrm{AR}$ denotes the autoregressive decoding of the discrete tokens, and $\mathrm{NAR}$ represents the non-autoregressive prediction of the continuous residuals. The operator $\parallel$ denotes sequence concatenation along the temporal dimension. The $\mathrm{Up}$ function aligns the temporal resolution of the generated discrete tokens $\hat{\mathbf{z}}_{q}$ with the continuous space using the defined up-sampling rate $r$. Finally, the $\mathrm{Decoder}$ module synthesizes the final audio waveform $\hat{\mathbf{s}}$ by combining both the discrete and continuous representations. The signed-log transform $f_\textrm{SLT}(x) = sign(x) \log (|x| + 1)$ \cite{slt} is used to improve training dynamics.

With a downsampled discrete track, the generation time is significantly reduced: $n_\text{cascade} = n_\text{full} / r + 1$, where $n_\text{cascade}$ is the number of Transformer steps required with cascading inference, $n_\text{full}$ is the number of full autoregressive steps, and $r = f_\textrm{base}/f_{model}$ is the scaling factor. For instance, generating a 10-second sample at $50~\text{Hz}$ traditionally takes 500 steps, but using a 12.5 Hz residual-enhanced model $500/4 + 1 = 126$ steps without the significant quality loss of the discrete-only model.

\section{Experimental Setup}

\begin{table*}
    \centering
   \caption{Comparison of downstream task (TTS/ASR) performance (best results in bold)
   }
\scriptsize

\begin{tabular}{ccccccccccc}
\toprule
& & \multicolumn{3}{c}{\textbf{TTS (generative task)}} & \multicolumn{2}{c}{\textbf{ASR (discriminative task)}}  & \\
\cmidrule(lr){3-5} \cmidrule(lr){6-7}
\textbf{Representation}    & \textbf{Frame rate (Hz)}&  \textbf{UTMOS} ($\uparrow$) &   \textbf{dWER} ($\downarrow$) & \textbf{SpkSim}  ($\uparrow$)   & \textbf{WER} ($\downarrow$) &   \textbf{CER} ($\downarrow$)   \\
\hline
Discrete-Only & 50.0  & 4.07 & 16.10 & 0.924 & 28.11 & 14.48 \\
\rowcolor{blue!6}
Hybrid (Ours) & 50.0 & \textbf{4.14} & \textbf{11.67} & \textbf{0.926} & \textbf{23.36} & \textbf{12.36} \\
\hline
 Discrete-Only  & 25.0  & 3.98 & \textbf{10.09} & 0.866 & 31.48 & 16.76 \\
 \rowcolor{blue!6}
 Hybrid (Ours) & 25.0  & \textbf{4.22} & 10.33 &\textbf{ 0.914} & \textbf{28.36} & \textbf{14.99} \\
\hline
Discrete-Only  & 12.5  &  1.99 & 32.97  & 0.853 & 28.50  & 14.19   \\
 \rowcolor{blue!6}
Hybrid (Ours) & 12.5  &  \textbf{4.10}  & \textbf{14.79} & \textbf{0.905} & \textbf{25.94} &   \textbf{12.86}   \\
 \hline
Discrete-Only & 6.25  & 1.44 &  121.00 &  0.707 & 29.13 & 15.45 \\
\rowcolor{blue!6}
Hybrid (Ours) & 6.25 & \textbf{3.08} & \textbf{48.00} & \textbf{0.834} & \textbf{27.36} & \textbf{13.62} \\
\bottomrule
\end{tabular}
    \label{tab:results-tts-asr}
\end{table*}

We use the 960-hour LibriTTS~\cite{libritts} dataset, an extension of LibriSpeech~\cite{librispeech} specifically optimized for TTS. While we train on both the \emph{clean} and \emph{other} (distorted) subsets for training, we strictly limit our evaluation to the \emph{clean} test set to maintain consistency. To align with the original FocalCodec setup and avoid out-of-distribution artifacts, we exclude any audio samples exceeding 20 seconds. During evaluation, ASR performance is computed on the full test set. For TTS, we uniformly sample a single subset of 1,000 utterances.  We implement our framework using the SpeechBrain~\cite{speechbrain_v1} toolkit. For reproducibility and to support the community, all source code and models will be made publicly available within the SpeechBrain project\footnote{\url{https://speechbrain.github.io/}}.

\subsection{Metrics}
We use the objective evaluation metrics listed below:

\begin{itemize}
    \item \textbf{Audio quality and naturalness:} We report \emph{UTMOS}~\cite{utmos} and \emph{NISQA}~\cite{nisqa}, neural estimators of the Mean Opinion Score (MOS, ranging from $1.0$ to $5.0$, where higher is better). While UTMOS evaluates perceived overall naturalness, NISQA specifically targets signal transmission quality.
    
    \item \textbf{Intelligibility:} We measure robustness against mispronunciations and acoustic artifacts using the differential Word Error Rate (\emph{dWER}). It is computed as the word-level edit distance between ASR transcriptions of the synthesized audio and the ground truth. Following standard benchmarks like DASB~\cite{dasb-benchmark}, we intentionally employ Whisper Small~\cite{whisper} with greedy decoding; a weaker ASR model is strictly preferable here, as it is less capable of implicitly compensating for underlying codec or synthesis flaws.
    
    \item \textbf{Speaker identity preservation:} We quantify voice fidelity via \emph{SpkSim} (ranging from $0.0$ to $1.0$, higher is better). This metric calculates the cosine similarity between latent embeddings extracted from a pretrained WavLM model fine-tuned for speaker verification (WavLM-SV), ensuring the synthesized vocal characteristics strongly match the original target.

    \item \textbf{Quantization efficiency:} To evaluate the effectiveness of discrete codebook use, we report \emph{Code Usage} (percentage of active codebook vectors) and \emph{Normalized Entropy} (token uniformity). High values indicate optimal vocabulary exploitation, preventing the index collapse typical of extreme low-bitrate compression.
\end{itemize}

\subsection{Tasks}
\label{tasks}

We evaluate our approach across three tasks. For TTS and ASR, input prompts and targets are combined into a single sequence using \texttt{[BOS]}, \texttt{[EOP]} (which separates the prompt from the generated target), and \texttt{[EOS]} control tokens.

\begin{enumerate}
    \item \textbf{Resynthesis:} Evaluates the standalone reconstruction quality of the WavLM-based HybridCodec by decoding ground-truth discrete tokens and continuous residuals directly through the vocoder, without a language model.
    
    \item \textbf{Text-To-Speech (TTS):} The model generates audio from a text prompt (\texttt{[BOS][Chars][EOP]}). The target consists of AR-generated discrete tokens followed by a NAR residual prediction pass. Quality is assessed via UTMOS, dWER, and SpkSim.
    
    \item \textbf{Automatic Speech Recognition (ASR):} The model maps audio prompts (discrete tokens and residuals) to text (\texttt{[Text][EOS]}). To isolate the impact of hybrid representations, we use simple greedy search, focusing on relative performance (Word Error Rate--WER and Character Error Rate--CER) rather than state-of-the-art benchmarking.
\end{enumerate}

\section{Results}
We first evaluate the reconstruction capabilities of our codec through resynthesis, before assessing its performance on downstream tasks. In both scenarios, we compare our hybrid approach against discrete-only baselines. 

\noindent\textbf{Resynthesis:} 
Table \ref{tab:results-resynth-wavlm} compares HybridCodec against state-of-the-art NACs \cite{dac, kyutai2024moshi, focalcodec, xin2024bigcodec}. To our knowledge, ours is the first approach to maintain such high semantic and speaker preservation at ultra-low frame rates (6.25 Hz). While baselines like FocalCodec \cite{focalcodec} degrade at lower frame rates, our approach remains remarkably robust and stable. At $12.5$\,Hz, HybridCodec achieves the best intelligibility with dWER of $1.47$, a significant improvement over the $7.94$ dWER of the discrete-only FocalCodec baseline. Speaker similarity also remains high (97.1) compared to other $12.5$\,Hz models like Mimi \cite{kyutai2024moshi} ($96.0$). Even at an extreme $6.25$\,Hz, performance remains nearly identical (3.98 UTMOS, 1.50 dWER). This shows that our residual information effectively mitigates the quantization penalty, offering an efficient alternative to high-frequency NACs.

\noindent\textbf{TTS:} Table \ref{tab:results-tts-asr} summarizes the results of discrete and hybrid TTS and ASR using HybridCodec within the HybridLM framework. Note that the discrete, non-finetuned version is identical to the publicly available, frame-rate-matched FocalCodec \cite{focalcodec}. For TTS, these results empirically validate our initial hypothesis: introducing a single non-autoregressive residual prediction step at the end of inference effectively mitigates the severe performance degradation typical of low-token-rate, discrete-only codecs. For instance, in the zero-shot setting at 12.5 Hz, our hybrid method more than doubles the UTMOS score ($4.10$ vs. $1.99$) and reduces the dWER by more than half ($14.79$ vs. $32.97$) compared to the discrete baseline. Similarly to our resynthesis findings, the performance gap between the hybrid and discrete representations widens as the operating frequency decreases. At an extremely low rate of $6.25$\,Hz, the hybrid approach still achieves a UTMOS of $3.08$ and a dWER of $48$, vastly outperforming the discrete-only representation ($1.44$ UTMOS and $121$ dWER). This demonstrates that as strict quantization discards acoustic details, our continuous refinement becomes important for recovering signal fidelity.

\noindent\textbf{ASR:} The inclusion of continuous residuals not only maintains semantic integrity but consistently improves transcription accuracy across all frame rates. While discrete tokens capture core semantics, continuous residuals provide acoustic cues that directly improve recognition. As shown in Table \ref{tab:results-tts-asr}, the hybrid approach lowers the WER at 50 Hz from 28.11 to 23.36, and the CER from 14.48 to 12.36. This performance gain holds even at higher compression levels: at 12.5 Hz, the hybrid model achieves a 25.94 WER (vs. 28.50 for the baseline), and at the extreme 6.25 Hz rate, it reduces the WER from 29.13 to 27.36. This is a critical success for our unified framework, confirming that the continuous acoustic information improves rather than interferes with the underlying semantic representations used by HybridLM.

\section{Conclusion}
This work introduced HybridCodec, a novel framework that bridges discrete efficiency and continuous acoustic fidelity at remarkably low frame rates. By combining discrete tokens with a non-autoregressive residual pathway, we recovered high-fidelity speech details at an ultra-low temporal resolution of 6.25\,Hz. Our results show that this hybrid approach outperforms a discrete-only baseline in TTS quality and intelligibility, while simultaneously reducing error rates in discriminative ASR tasks.
The HybridLM architecture further shows that these dual representations can be unified within a single Transformer via AdaLN. By operating at such extreme compression levels, our method significantly reduces the number of inference steps, offering a highly efficient alternative for long-form synthesis. \hide{Future work will focus on generalizing this paradigm to other speech tasks and exploring it as a simpler, more efficient alternative to multi-codebook residual vector quantization (RVQ)~\cite{rvq}.}

\newpage
\section{Generative AI Use Disclosure}
LLMs~\cite{chatgpt, copilot, opus, gemini, ai2_asta} have been used for advanced search, for boilerplate automation, and as a technical reference. LLMs have not been used to author text for the paper, except BibTeX formatting and grammar/wording revisions. LLM outputs were manually reviewed.

\section{Acknowledgments}
We gratefully acknowledge the support of NSERC, the Digital Research Alliance of Canada (alliancecan.ca), Translated (Imminent Program), and Apple (Seed Grant) through research funding, computing resources, and donations. Samir Sadok was supported by the VisaSpeech Inria Associated Team initiative.

\bibliographystyle{IEEEtran}
\bibliography{mybib}

\end{document}